\documentclass{WileyMSP-template}
\usepackage{amsmath}
\usepackage{xcolor}
\usepackage{multirow}
\usepackage{rotating}
\usepackage[font=sf,labelfont=bf]{caption}

\begin{document}

\pagestyle{fancy}
\rhead{\includegraphics[width=2.5cm]{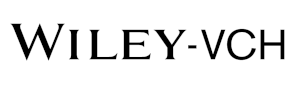}}

\title{Hyper-Yoshimura: How a Small Modification to a Classical Origami Pattern Enables Sequential Buckling, Self-Packing, and Autonomous Reconfiguration}

\maketitle


\author{Ziyang Zhou$^{1\dagger}$}
\author{Yogesh Phalak$^{1\dagger}$}
\author{Vishrut Deshpande$^{1}$,}
\author{Ian Walker$^{2}$}
\author{Suyi Li$^{1\ast}$}

\begin{affiliations}
$^{1}$Department of Mechanical Engineering, \\Virginia Tech, Blacksburg \& 24060, USA.\\[6pt]
$^{2}$Department of Electrical Engineering and Computer Science, \\University of Wyoming,  Laramie \& 82071, USA. \\[6pt]
$^\ast$Correspondent email address: \texttt{suyili@vt.edu} \\[6pt]
$^\dagger$These authors contributed equally to this work.

\end{affiliations}


\keywords{\\Yoshimura origami, elastic instabilities, sequential buckling,\\ self-packing, deployable robotics}

\begin{abstract}

Buckling instabilities can enable rapid shape transformation, programmable mechanical responses, and many other engineering functions. Yet programming and controlling buckling remain a challenge. This study explores origami as a geometric framework to translate thin-shell buckling into functionality.  Specifically, this study introduces hyper-Yoshimura, an extension of a classical origami design, to enable sequenced buckling and autonomous reconfiguration. Kinematic and energetic analyses show that, by allowing the sector angle $\beta$ beyond the flat-foldability condition (i.e., $90^\circ/n<\beta<45^\circ$), hyper-Yoshimura obtains a self-\textbf{P}acked state alongside the \textbf{D}eployed and \textbf{F}olded states.  Moreover, this sector angle also dictates the critical buckling force required to switch the hyper-Yoshimura from Deployed to Folded states. Therefore, a hybrid design with different $\beta$ angles, along with embedded pneumatic and tendon actuation, allows a 3D-printed Yoshimura boom to repeatedly buckle, deploy, and switch among many configurations on demand. The engineering functions of hyper-Yoshimura are demonstrated through object manipulation with an expandable workspace, re-positioning and re-orienting solar panels mounted at its tip, and carrying a payload 28.5 times its structural weight. These results establish hyper-Yoshimura as a new design language for engineering buckling instability and its applications in cylindrical shell structures.

\end{abstract}

\section{Introduction}
Buckling---an elastic instability that occurs when a structure is compressed beyond its critical load---has traditionally been regarded as a failure mode because it can introduce abrupt and substantial degradation in shape, stiffness, or load-carrying capacity. However, if buckling is intentionally controlled, it can become an engineering framework for creating unique functions, including rapid shape transformations \cite{bertoldi2017flexible, kong2026origami,boron2026stochastic}, multiple stable configurations \cite{yan2025snap, zhou2025multistable}, structural stiffness modulation \cite{levine2021materials, gao2022stiffness}, and mechanical memory \cite{jiao2023mechanical, chen2021reprogrammable}. These functions are increasingly used in mechanical metamaterials \cite{zaiser2023disordered, sun2026magnetic}, soft robotics \cite{zhou2024self, luo2024intrinsically}, morphing and deployable structures \cite{djellouli2024shell, bichara2023multi}, flexible electronics \cite{yang2024kirigami}, adaptive energy systems \cite{lamoureux2015dynamic,wang2026coupled}, and physical computing \cite{wang2026embodying}. Consequently, our focus has shifted from merely avoiding buckling to strategically harnessing it, and researchers have been seeking effective strategies to control the onset of buckling, prescribe the resulting deformation modes, and exploit the ensuing nonlinear behaviors to advance structural performance and functionality.

\begin{figure}[b!]
    \centering
    \includegraphics[scale=1.0]{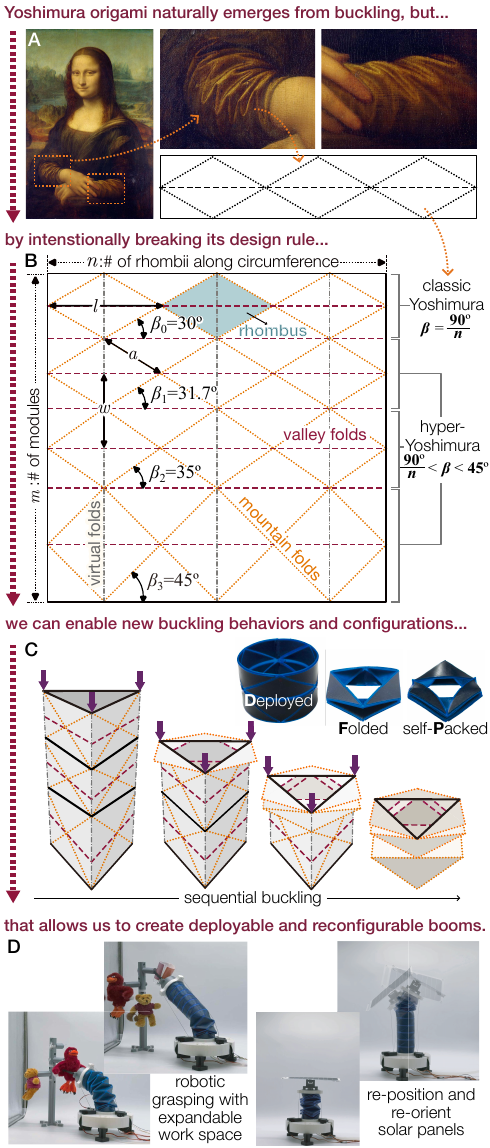}
    \caption{\textbf{Motivation and Overview.} (A) As Lady Mona Lisa rolled up her sleeve, the fabric inevitably buckled into a Yoshimura-like origami pattern, a deformation that Leonardo da Vinci captured with remarkable accuracy in the painting. (B) A four-module Yoshimura pattern consisting of the classical design ($\beta=90^\circ/n$) and the new, ``hyper'' design proposed in this study ($90^\circ/n\leq\beta\leq45^\circ$).  (C) A hyper-Yoshimura with different $\beta$ values exhibiting sequential buckling, and the distinct \textbf{D}eployed, \textbf{F}olded, and self-\textbf{P}acked states. (D) Engineering demonstrations in this study include a robotic gripper with an expandable workspace and a deployable boom that can reposition and reorient solar panels.}
    \label{fig:overview}
\end{figure}

\medskip
In this regard, origami provides a unified geometric framework for organizing buckling deformations in thin sheets and shells \cite{lu2023origami, misseroni2024origami}. Beyond its origins as an ancient art form shaped by human creativity, origami can also emerge naturally from buckling instabilities. For example, biaxial in-plane compression of a flat sheet can produce the Miura-ori pattern~\cite{schenk2013geometry, wang2023building}; axial compression of a thin cylindrical shell can generate the Yoshimura pattern~\cite{deshpande2024golden} (\textbf{Figure~\ref{fig:overview}}A); and superimposing torsion on axial compression gives rise to the Kresling pattern~\cite{zhang2023plug, ze2022soft}. In this sense, these origami patterns can be viewed as being ``designed'' by the physics of buckling instability.

\medskip
This intriguing connection between origami and buckling motivates the central question of this study: If origami can emerge from the buckling of thin shells, can we reverse the process?  Can we customize an origami pattern to \textit{program} how a thin shell structure buckles? More importantly, can such geometric customization give rise to buckling behaviors that would otherwise be inaccessible through natural buckling alone?

\medskip
Here, we use the cylindrical Yoshimura pattern and its associated axial buckling mode as a canonical example to answer these questions. The Yoshimura pattern consists of a tessellation of triangular facets (or rhombi), whose geometry is fully defined by two parameters: the sector angle $\beta$, which determines the rhombus shape, and $n$, the number of rhombi around the cylinder’s circumference (Figure \ref{fig:overview}B). In the classical Yoshimura pattern, these parameters satisfy the flat-foldability condition: $\beta=90^\circ/n$. This seemingly simple geometric relationship embodies an important physical reality: When a cylindrical shell buckles under axial compression, it naturally develops creases and folds that allow it to collapse into a flat ring. Consequently, the classical, flat-foldable Yoshimura pattern corresponds to one of the most energetically favorable post-buckling deformation paths for a cylindrical shell while preserving axisymmetry.

\medskip
However, in this study, we intentionally violate and relax this design rule by allowing $\beta>90^\circ/n$ (Figure \ref{fig:overview}B). Such larger sector angles increase the critical compressive load required to initiate buckling and therefore do not emerge naturally. \textit{However, they fundamentally alter the post-buckling behavior in two important ways}. First, instead of collapsing into a flat ring, the buckled cylinder transforms into a wavy geometry that resembles a hyperbolic surface. We therefore refer to this generalized Yoshimura pattern as the ``hyper-Yoshimura.'' Remarkably, this collapsed hyperbolic configuration is elastically stable, enabling the buckled cylindrical shell to \textit{self-pack} into a compact volume. As a result, the hyper-Yoshimura can access a stiff \textbf{D}eployed state (before buckling), a compliant \textbf{F}olded state (post-buckling), and a compact self-\textbf{P}acked state (after collapsing). We referred to them as the ``\textbf{PDF}'' states for brevity ({Figure \ref{fig:overview}}C). Second, different $\beta$ angles can be seamlessly integrated within a single Yoshimura pattern to sequence buckling events. This sequential buckling enables a pre-programmed cascade of localized snap-through instabilities under a single global compressive load. When adjacent modules possess sufficiently different buckling thresholds, the resulting transitions produce multistage shape transformations, stabilize useful intermediate configurations, and tailor the overall force–displacement response. 

\medskip
Moreover, by combining multi-material 3D printing with integrated pneumatic–tendon actuation, we render the sequential buckling process \textit{reversible} and \textit{repeatable}. This enables the hyper-Yoshimura to autonomously reconfigure among different combinations of \textbf{PDF} states. 
To demonstrate the application potential of this capability, and inspired by extraterrestrial habitats, we developed a six-module deployable hyper-Yoshimura crane that manipulates objects within an expandable workspace, manipulates objects, repositions and reorients a solar panel mounted at its tip, and supports payloads without requiring continuous pneumatic pressurization (Figure~\ref{fig:overview}D).

\medskip
Together, these results show how a small geometric modification to a classical buckling-derived origami pattern transforms instability from an unpredictable failure mode into an engineering tool for creating self-packing, sequential deployment, and autonomous robotic reconfigurations.

\medskip


\section{The Emergence of Deployed, Folded, and Self-Packed States}

Here, we first establish the geometric design rules of the hyper-Yoshimura. We then show that when $\frac{90^\circ}{n} < \beta < 45^\circ$, the \textbf{D}eployed, \textbf{F}olded, and self-\textbf{P}acked states constitute three \textit{kinematically admissible} configurations (i.e., no material stretching), and all these states are accessible through compressive loading.

\subsection*{The Design Rules and Geometric Parametrization of hyper-Yoshimura} \label{subsec:GeometricFoundations}


The Yoshimura origami pattern consists of a periodic tessellation of mountain and valley folds that define an array of fundamental rhombus shapes (Figure~\ref{fig:overview}B). The valley folds have length $l$, the mountain folds have length $a$, and the distance between adjacent valley folds is $w$. The tessellation is further parameterized by two integers: $n$, the number of rhombi along the circumferential direction, and $m$, the number of rhombi layers along the height. After folding, each layer of rhombi forms a \emph{module}; accordingly, $m$ can also be interpreted as the number of axially stacked modules. The most critical design parameter governing the tessellation’s geometry is the sector angle $\beta$, which defines the angular separation between mountain and valley folds within the plane. 
 
\medskip
To non-dimensionalize the geometry, we introduce a reference length $L$ such that $l/L = 1$, so all geometric quantities are normalized relative to the valley fold length.  This way, once $n$ and $\beta$ are specified, the Yoshimura tessellation geometry is uniquely determined, and quantities such as $w$ and $a$ can be derived consistently based on these fundamental parameters. For example, $w = \tan{\beta}$, and $a=(2\cos\beta)^{-1}$.

\begin{figure}[b!]
    \centering
    \includegraphics[scale=1]{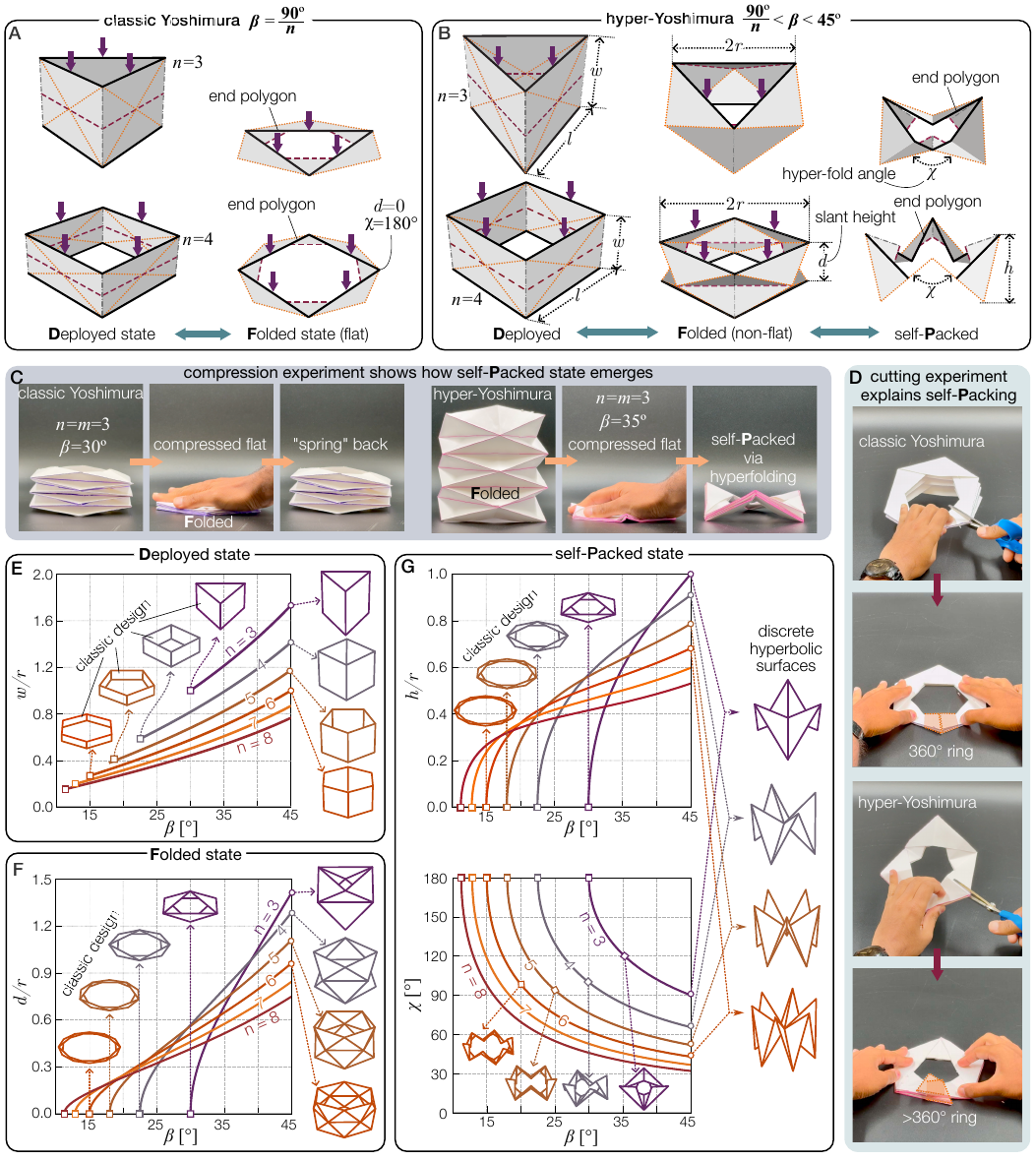}
    \caption{\textbf{The Geometry of hyper-Yoshimura and its Kinematically Admissible States.} 
    (A) 3D renderings of $n=3$ and $n=4$ modules featuring the classical Yoshimura design. Here, their \textbf{F}olded state are exactly flat. 
    (B) 3D renderings of $n=3$ and $n=4$ modules featuring hyper-Yoshimura design, showing their distinct \textbf{D}eployed, \textbf{F}olded, and self-\textbf{P}acked states. 
    (C) Comparison between classical Yoshimura and hyper-Yoshimura under axial compression. The self-packing behavior allows the hyper-Yoshimura to settle into a compact, discrete hyperbolic surface without ``springing back''. 
    (D) Cutting experiment revealing the residual over-curvature associated with the self-Packing.
    (E-F) Correlations between the sector angle $\beta$ and the module's height at the \textbf{D}eployed-, \textbf{F}olded-, and self-\textbf{P}acked states ($w$, $d$, and $h$, respectively). The lower plot in (G) also shows that $\chi$ begins at $180^\circ$ at the classical, flat-foldability boundary and decreases with $\beta$, corresponding to increasing hyper-fold. In these plots, square symbols correspond to the lower limit of $\beta=90^\circ/n$ (classic Yoshimura), whereas the circular symbols correponds to the upper limit of $\beta=45^\circ$.}
    \label{fig:states}
\end{figure}

\medskip
Upon folding along the mountain and valley creases and connecting the opposite edges, the Yoshimura tessellation transforms into its classical three-dimensional shape: a column-like structure with a zigzagged surface, where each rhombus becomes a local valley (\textbf{Figure~\ref{fig:states}}A,B). We refer to this configuration as the \emph{\textbf{F}olded state} (\textbf{F}) hereafter. In this state, the top and bottom boundaries of each module form an $n$-sided polygon. Notably, the side length of these polygons remains identical to the unfolded valley fold length $l$. 
Two critical geometric variables emerge in the \textbf{F}olded state: $\theta$, the dihedral angle between adjacent facets, and $d$, the distance between the centroids of top and bottom boundary polygons of a module (Figure~\ref{fig:states}b). This slant height $d$ directly measures the module's deformation during bucking and reconfiguration, and thus plays a pivotal role in the subsequent analysis. 

\medskip
From the \textbf{F}olded state, we can stretch the Yoshimura by pushing the rhombus facets outwards, creating a load-bearing boom. We refer to this configuration as the \emph{\textbf{D}eployed state} (\textbf{D}). In this state, the original mountain and valley creases are no longer folded, so $\theta=180^\circ$. Indeed, transforming Yoshimura from \textbf{D} to \textbf{F} states exactly represents the compression buckling of a cylindrical shell. In this study, we idealize the \textbf{D}eployed state by introducing \emph{virtual folds} so it forms a symmetric and regular $n$-polygonal column. 

\medskip
Together, $\theta$ and $d$ describe the shape transformation of Yoshimura at the \textbf{F}olded and \textbf{D}eployed states, so they are called the \emph{transformation variables} (as opposed to the \emph{design variables} $\beta$, $n$, and $m$). By systematically defining the relationships among these variables, Yoshimura can be studied as a parametric structure in which changes in crease geometry produce predictable changes in 3D kinematics. This parametric framework provides the basis for the kinematic analysis below.

\subsection*{Kinematic Transformation From Deployed (\textbf{D}) to Folded (\textbf{F}) State} 


\medskip
The module starts in the \textbf{D}eployed state (\textbf{D}), where the slant height reaches its maximum value $d = w$, and the module's top and bottom boundaries form concentric and regular $n$-gons (Figure~\ref{fig:states}E).  All creases retain their original lengths, ensuring this configuration remains kinematically admissible.
From this state, one can fold the module by decreasing $\theta$ while preserving the symmetry between the top and bottom boundary $n$-gons. As $\theta$ decreases, the slant height $d$ also reduces, and the structure approaches a second kinematically admissible state: the \textbf{F}olded state (\textbf{F}). 


\medskip
Kinematic admissibility of this \textbf{F}olded state requires the facets to tilt in a manner that satisfies two geometric constraints arising from axial projection and circumferential closure:

\begin{align}
    d &= w \sin{\left(\frac{\theta}{2}\right)}, \\
    w \cos{\left(\frac{\theta}{2}\right)} &= \tan\left(\frac{90^\circ}{n}\right).
    \label{eq:d_folded}
\end{align}

Together, these equations correlate the transformation variables $d$ and $\theta$ to the design variables $\beta$ and $n$. Using the non-dimensionalization $w=\tan\beta$, they give
\begin{align}
    \label{eq:folded_state_d}
    d_\mathbf{F} &= \sqrt{\tan^2\!\beta-\tan^2\!\left(\frac{90^\circ}{n}\right)}, \\
    \theta_\mathbf{F} &= 2\sin^{-1}\!\left(\frac{d_\mathrm{F}}{w}\right)
    =2\cos^{-1}\!\left[\frac{\tan(90^\circ/n)}{\tan\beta}\right].
    \label{eq:folded_state_theta}
\end{align}

Note that the inverse-trigonometric relation follows directly from $d=w\sin(\theta/2)$ and retains the full dihedral-angle convention, so $\theta=180^\circ$ in the Deployed state. For real solution to exist for Eq. (\ref{eq:folded_state_d}) and (\ref{eq:folded_state_theta}), $\beta$ must satify the following condition:
\begin{equation}
    \beta \geq \frac{90^\circ}{n}.
\end{equation}

The limiting case, $\beta = 90^\circ/n$, corresponds to the classical Yoshimura design. Under this condition, the \textbf{F}olded state collapses completely into a flat shape with $d=0$, and all module boundaries converge into a single plane (i.e., flat-foldability, Figure~\ref{fig:states}F). Note that, in physical prototyping, the classical Yoshimura springs back slightly from this flat-folded shape because of crease bending stiffness and finite material thickness (Figure~\ref{fig:states}C and Movie S1). 

%

\subsection*{Kinematic Transformation From Folded (F) to self-Packed (P) State}


When $\beta>\frac{90^\circ}{n}$, the \textbf{F}olded Yoshimura has non-zero height, but we can compress it further by twisting the module's boundary vertices out of plane in an alternating, up-and-down fashion and folding the triangular facets along the virtual folds. The resulting compact ``wavy'' surface resembles a discretized hyperbolic surface and is illustrated for $n=3$ and $n=4$ in Figure~\ref{fig:states}(A, B). 
We refer to this process as ``hyper-folding'' and define the ``hyper-fold angle'' $\chi$ as the dihedral angle between the two triangular facets adjacent to a virtual crease. Thus, $\chi=180^\circ$ when these facets are locally coplanar and decreases as they hyper-fold. (The complementary rotation away from coplanarity is $180^\circ-\chi$; throughout this work, $\chi$ refers only to the included dihedral angle.) This angle is expressed in terms of the design parameters $\beta$ and $n$ as
\begin{equation}
    \cos{\chi} = 1-\frac{2\sin^2{\left( \frac{90^\circ}{n} \right)}}{\sin^2{\beta}}.
\end{equation}

The departure of $\chi$ from $180^\circ$ quantifies the local folding required for compressing Yoshimura beyond the classical flat-foldability condition. As shown in the lower plot of Figure~\ref{fig:states}(G), $\chi$ decreases with $\beta$. For $n=3$, it decreases from $180^\circ$ at $\beta=30^\circ$ to approximately $121.3^\circ$ at $\beta=35^\circ$ and approaches $90^\circ$ as $\beta\to45^\circ$. Accordingly, a smaller included angle represents stronger folding across the virtual crease.
Furthermore, the inner radial position of the folds --- measured as the minimum radial distance from the origin to the newly formed twisted layer --- is given by:
\begin{equation}
    r_\text{in} = \frac{1}{2\tan(2\beta)}
\end{equation}

This function indicates a critical kinematic constraint: $ r_\text{in} \to 0 $ as $\beta \to 45^\circ$.  When $\beta > 45^\circ$, $r$ becomes negative, indicating self-intersections and thus geometric infeasibility. This upper bound is reflected in the geometric renders in Figure~\ref{fig:states}(G). Therefore, to ensure kinematic admissibility of \textbf{D}, \textbf{F}, and \textbf{P}, $\beta$ must lie strictly within the range:
\begin{equation}
    \frac{90^\circ}{n} < \beta < 45^\circ.
    \label{eq-beta}
\end{equation}

We designate this entire class of Yoshimura with the extended $\beta$ angle range as ``Hyper-Yoshimura.'' A complete summary of the parametric bounds and geometric conditions is provided in Table~\ref{tab:states_parameters}. The formulations above also yield the connection between the sector angle $\beta$ and the transformation variables, summarized in Figure~\ref{fig:states}(E-G): the deployed height $w$, folded height $d$, packed height $h$, and hyper-fold angle $\chi$ together define the 3D shapes of hyper-Yoshimura at \textbf{D}, \textbf{F}, and \textbf{P} states (more details in SI Sections 1 and 2 regarding hyper-Yoshimura with higher $n$).

\medskip
Although the hyper-folded self-\textbf{P}acked state is geometrically admissible, its stability cannot be established from kinematics alone. Experimentally, the compressed hyper-Yoshimura retains a distinctly compact configuration after removal of the external load, whereas the classical flat-foldable Yoshimura partially recovers upon unloading (Figure~\ref{fig:states}C and Movie S1). A cutting experiment provides further insight. When the self-packed hyper-Yoshimura is cut along a virtual crease, the released structure opens into an $n$-polygonal ring whose accumulated circumferential angle exceeds $360^\circ$. In contrast, the classical Yoshimura yields a ring with an accumulated angle of exactly $360^\circ$ (Figure~\ref{fig:states}D and Movie S1). This excess circumferential angle leads to residual over-curvature in the hyper-folded geometry; therefore, geometric frustration stabilizes the self-\textbf{P}acked state's wavy configuration.


\section{Analyzing the Buckling Mechanics of hyper-Yoshimura}
\label{sec:model}
%
%
Although the geometric construction above establishes the kinematic admissibility of \textbf{D}, \textbf{F}, and \textbf{P} states, buckling behaviors are governed not only by kinematics but also by the underlying mechanics. Therefore, one must quantify both the energy barrier separating neighboring states and the peak reaction force under prescribed axial compression. The former describes the structure’s potential energy landscape, whereas the latter determines the critical load required to trigger buckling. Accordingly, we evaluate the potential energy landscape and the axial force--compression response of a single hyper-Yoshimura module. The results, as summarized in \textbf{Figure~\ref{fig:single_module}}, show that $\beta$ independently regulates the energy barrier and the critical buckling force, which subsequently govern the sequential buckling behavior of stacked hyper-Yoshimura booms.

\subsection*{Mechanics Modeling using Discrete Bar-network}
\label{subsec:kinematics}

To elucidate the buckling mechanics of the hyper-Yoshimura without unnecessary complexity, we restrict our analysis to the axisymmetric deformation of a single module with $n=3$, for which the sector angle $\beta$ must lie within $[30^\circ \; 45^\circ]$ according to Eq.~(\ref{eq-beta}). Under this assumption, the hyper-Yoshimura module's deformation can be described by a single scalar variable: the slant height $d$. We further define a complementary measure, the normalized ``compression ratio,'' as $c = 1 - d/\tan\beta$, which varies from $c=0$ in the \textbf{D}eployed state to $c=1$ in the fully compressed self-\textbf{P}acked state. Based on this parametrization, we employ a \textit{discrete bar-network model} to compute the elastic energy of the $n=3$ hyper-Yoshimura module as $c$ varies from 0 to 1 and $\beta$ ranges from $30^\circ$ to $45^\circ$.

\begin{figure}[b!]
    \centering
    \includegraphics[width=\textwidth]{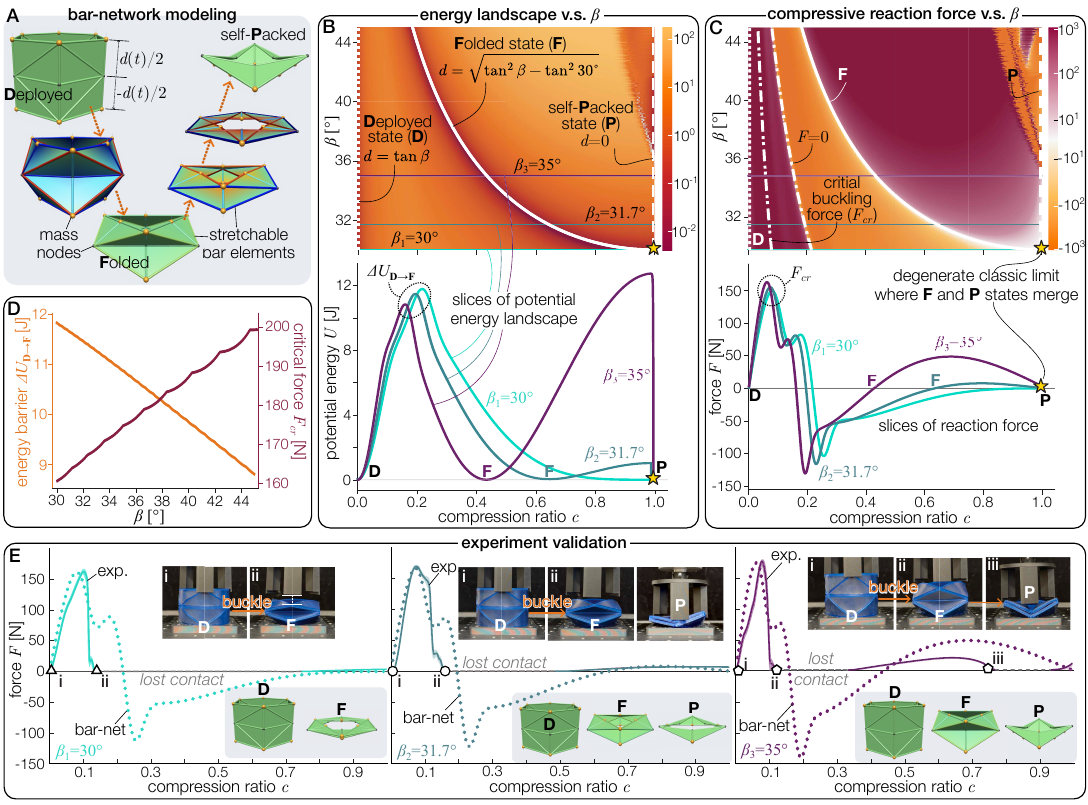}
        \caption{\textbf{Mechancis Analysis of the hyper-Yoshimura's Buckling.} 
        (A) Snapshots from a bar-network simulation illustrating how elastic stretching/compression of the bar elements can capture the overall potential-energy landscape.
        (B) Potential-energy landscape, $U$, of the hyper-Yoshimura modules over the design--configuration space. The loci of the \textbf{D}eployed, \textbf{F}olded, and self-\textbf{P}acked states, obtained from kinematic analysis, coincide precisely with the simulated energy minima. The lower subplot presents representative energy-landscape slices for $\beta=30^\circ$, $31.72^\circ$ (the golden-ratio angle), and $35^\circ$. Yellow stars indicate the degenerate classical endpoint at $(\beta,c)=(30^\circ,1)$, where the \textbf{F}olded and self-\textbf{P}acked configurations coincide.
        (C) Compressive reaction force, $F$, of the hyper-Yoshimura modules over the design--configuration space. The lower subplot shows representative force--compression curves for the same sector angles.
        (D) Dependence of the potential-energy barrier, $\Delta U_{\mathbf{D}\rightarrow\mathbf{F}}$, and critical buckling force, $F_\mathrm{cr}$, on the sector angle $\beta$. Such correlation is the basis of sequential buckling discussed later.
        (E) Comparison between experimental measurements and bar-network simulations for three 3D-printed hyper-Yoshimura modules. Here, $i$ denotes the initial \textbf{D}eployed state at the beginning of the compression loading cycle; $ii$ marks the onset of buckling toward the \textbf{F}olded state, after which the samples lose contact with the load cell; and $iii$ denotes the transition to the self-\textbf{P}acked state. Overall, the simulations accurately capture the relationship between the buckling response and the underlying Yoshimura crease geometry.(Movie S2)}
    \label{fig:single_module}
\end{figure}

\medskip
The bar-network model uses stretchable bar elements to represent the mountain, valley, and virtual folds in the hyper-Yoshimura (Figure~\ref{fig:single_module}A) and connects these bar elements with mass nodes, which also serve as universal joints. The bar elements' rest lengths are prescribed according to the geometric design rules; therefore, the bar network is stress-free in the kinematically admissible \textbf{D}, \textbf{F}, and \textbf{P} states, and any deviation from these states induces elastic deformation in the bars. This framework enables us to probe the energy landscape and reaction force of the hyper-Yoshimura by varying the slant height $d$ between its top and bottom end polygons (Methods).

\medskip
More specifically, the six nodes on the top polygon are prescribed with a common height of $+d/2$ along the $z$-axis, while the six nodes on the bottom polygon share a common height of $-d/2$. Their displacements along the $x$- and $y$-axes are unconstrained, and all other nodes are fully free.
Based on this setup, we simulate a complete compression cycle by prescribing $d$ along a smooth loading path, $d(t): \tan\beta \rightarrow 0$ (equivalently, $c: 0 \rightarrow 1$), corresponding to a displacement-controlled test. 
%
%
We set the damping ratio $\zeta$ large enough to ensure quasi-static simulations; that is, each instantaneous configuration at a prescribed slant height $d(t)$ corresponds to the minimum-energy state of the bar network. The instantaneous elastic energy $U(t)$ and the prescribed coordinate $d(t)$ therefore define the energy landscape $U(d;\beta)$, from which the axial force, $ F(d;\beta)=-\partial U/\partial d$, is obtained by numerical differentiation.

\subsection*{The Design--Configuration space of hyper-Yoshimura \label{subsec:config_space}}
\medskip
We simulate the compression loading cycles on many hyper-Yoshimura designs over the admissible range of sector angles, $30^\circ < \beta < 45^\circ$, and compile the resulting potential-energy landscapes and axial-force responses into a two-dimensional ``design--configuration space,'' denoted by $U(\beta,c)$ and $F(\beta,c)$ (Figure~\ref{fig:single_module}B,C). These representations show how hyper-Yoshimura's buckling relates to the underlying crease geometry.
Let $c_\mathrm{cr}(\beta)$ denote the elastic instability point between the \textbf{D}eployed and \textbf{F}olded states, i.e., the intervening local maximum in $U$. The potential-energy barrier and the critical buckling force associated with the transition from \textbf{D}eployed to \textbf{F}olded state are then given by
\begin{equation}
\begin{aligned}
    \Delta U_{\mathbf{D}\rightarrow \mathbf{F}}(\beta) &= U\!\left(\beta,c_\mathrm{cr}\right)-U\!\left(\beta,c_\mathbf{D}\right),\\
    F_\mathrm{cr}(\beta) &= \max_{c_\mathbf{D}\leq c\leq c_\mathrm{cr}} F(\beta,c),
\end{aligned}
\label{eq:barrier_force}
\end{equation}
where $c_\mathbf{D}=0$ corresponds to zero compression in the \textbf{D}eployed state. Here, we take the \textbf{D}eployed state as the zero-energy reference, so $\Delta U_{\mathbf{D}\rightarrow \mathbf{F}}$ is simply the energy of the intervening barrier.
The results for $\beta=30^\circ$ provide a useful boundary check: At this limit, the \textbf{F}olded and self-\textbf{P}acked states coincide at $c=1$ (yellow star in Figure~\ref{fig:single_module}B, C). Once $\beta>30^\circ$, the locus of \textbf{F}olded states shifts to $c<1$, whereas the self-\textbf{P}acked locus remains fixed at $c=1$. Figure~\ref{fig:single_module}C plots $\Delta U_{\mathbf{D}\rightarrow \mathbf{F}}$ and $F_\mathrm{cr}$ directly as functions of $\beta$, showing that the former decreases with increasing $\beta$, whereas the latter increases. Such opposite trends are not contradictory. Rather, together with the representative slices of the potential-energy landscape shown in Figure~\ref{fig:single_module}(B, C), they indicate that the barrier becomes both narrower and steeper as $\beta$ increases. Consequently, a smaller energy barrier separates the \textbf{D}eployed and \textbf{F}olded states, yet a larger critical compressive force is required to trigger buckling. 

\medskip
To validate the correlation between buckling and $\beta$ design, we compare the predictions of the bar-network model with compression-buckling experiments performed on 3D-printed single-module hyper-Yoshimura samples with $\beta=30^\circ$, $31.72^\circ$, and $35^\circ$ (Instron 6800 Series with a 1KN load cell; Figure~\ref{fig:single_module}E; more fabrication details in Methods). Note that $\beta=31.72^\circ$ is the Golden Ratio angle highlighted in a previous study~\cite{deshpande2024golden}.  The samples have thin, flexible TPU95A in the crease regions and relatively stiff Nylon in the facets (Ultimaker S5 FDM). For all three sector angles, the simulation and experimental results show the same qualitative behavior: the compressive force increases from the \textbf{D}eployed state to a well-defined maximum, and then decreases as the hyper-Yoshimura buckles into the \textbf{F}olded state. The predicted values of $c_\mathrm{cr}$ and $F_\mathrm{cr}$ agree quantitatively with the measurements, indicating that the model captures the critical force threshold governing the onset of buckling (Movie S2).

\medskip
Beyond the critical point, the measured force drops to zero because the sample loses contact with the load cell, whereas the model continues to follow the post-buckling branch; this loss of contact is indicated by the dashed gray sections in the experimental curves. For the $\beta=35^\circ$ sample, the structure re-establishes contact with the load cell upon further compression and subsequently follows a second positive-force branch before reaching the self-\textbf{P}acked configuration. 

\medskip
Notably, at maximum compression ($c=1$), both the $\beta=31.72^\circ$ and $\beta=35^\circ$ samples exhibit the characteristic hyperbolic surface geometry of the self-\textbf{P}acked state. However, self-packing in the $\beta=31.7^\circ$ sample is comparatively weak because of the finite thickness of the 3D-printed material. Consequently, this sample springs back to the \textbf{F}olded state when the compressive load is removed, whereas the $\beta=35^\circ$ sample remains stable in the self-\textbf{P}acked state. This behavior explains one factor behind the discrepancy between the measured and simulated post-buckling responses; additional contributions may arise from self-contact and fabrication imperfections.


\section{Sequencing the Buckling with Hybrid Designs}

\begin{figure}[h!]
    \centering
    \includegraphics[scale=1.0]{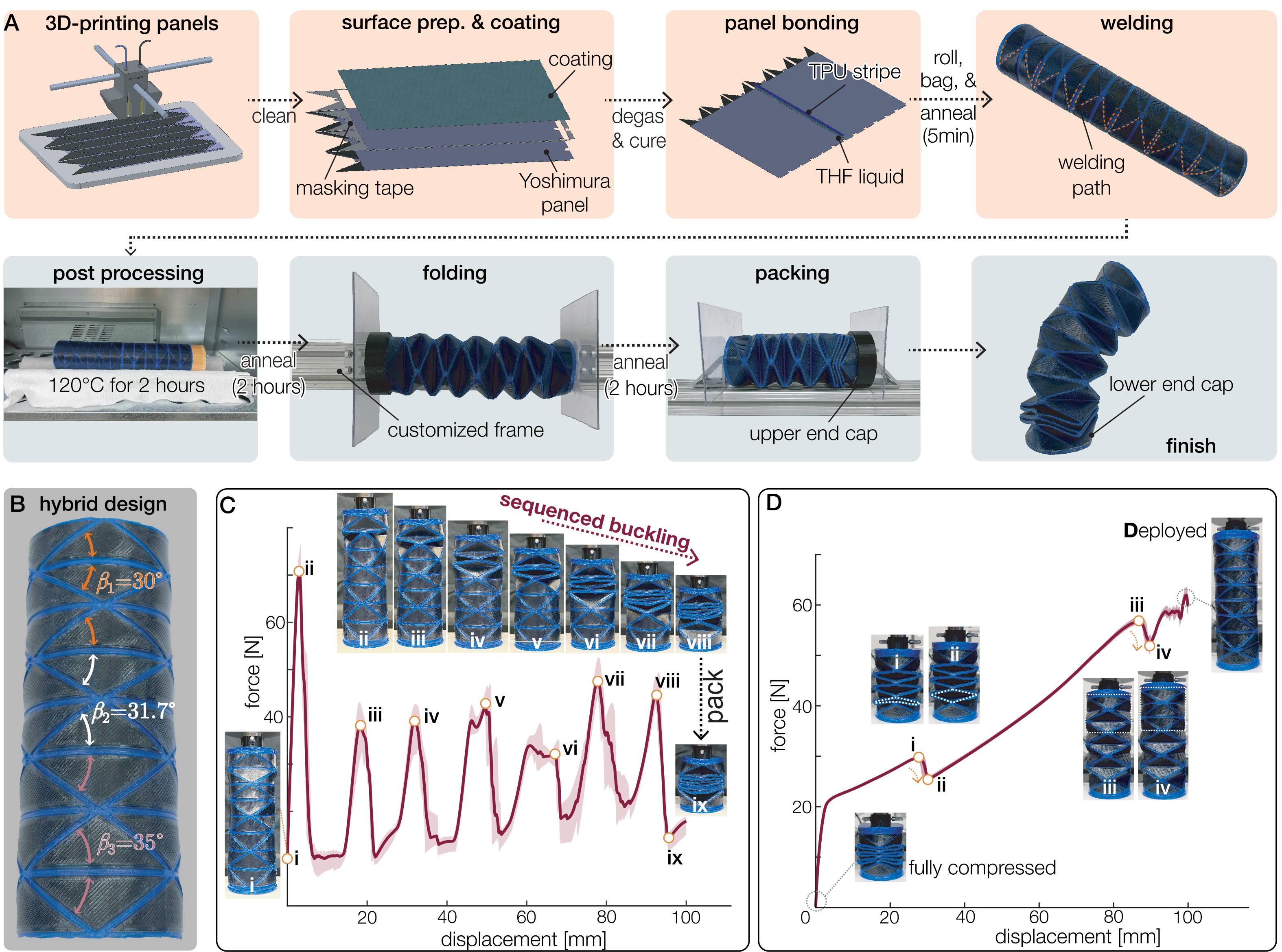}
    \caption{\textbf{Sequence Local Buckling with Hybrid Designs of Different $\beta$ Angles.} (A) Simplified fabrication procedure of the hyper-Yoshimura specimen. Note that annealing is conducted at $140^\circ$C. (B) Finished prototype, highlighting the different sector angles. (C, D) Mechanical compression and extension test results, respectively. The solid lines are the averaged data from 7 successive loading cycles, and the shaded bands are the standard deviation.}
    \label{fig:hybrid}
\end{figure}

The results above on single hyper-Yoshimura modules suggest a direct strategy for programming and controlling the buckling response of a hollow boom structure: one can assemble Yoshimura modules with different $\beta$ values so that their buckling thresholds are sequenced deterministically. To test this strategy, we fabricate a \textit{hybrid} Yoshimura sample consisting of three modules ($n=3$, $m=3$) with $\beta=30^\circ$, $31.72^\circ$, and $35^\circ$ arranged from the top to bottom. This spatial variation in $\beta$ is shown schematically in Figure~\ref{fig:overview}B and implemented in the 3D-printed specimen shown in \textbf{Figure}~\ref{fig:hybrid}(A, B) (Methods). All other geometric parameters, including facet dimensions and material thickness, are held constant across the three modules to isolate the effect of sector angle on mechanical response. In addition, we append half-length Yoshimura modules to both ends of the sample, allowing end caps to be installed to constrain boundary deformation during testing.


\medskip
Figure~\ref{fig:hybrid}(C) shows the force--displacement response of the hybrid Yoshimura sample under quasi-static compression. The reaction force curve exhibits a sequence of peaks and drops, each corresponding to a local buckling event when a layer of rhombus facets switches from \textbf{D}eployed to \textbf{F}olded states. Consistent with the $F_\mathrm{cr}(\beta)$ trend in Figure~\ref{fig:single_module}(D), the module with the smallest $\beta$ has the lowest critical force and therefore buckles first. The buckling sequence progresses from top to bottom, with each pop-in event marked by a sharp drop in reaction force (points $i$ to $viii$), eventually creating a ``cascading collapse''. The final drop between points $viii$ and $ix$ is different: it correlates with self-packing of the bottom $\beta=35^\circ$ module.

\medskip
In addition, we conduct a reversed tensile test by stretching the Yoshimura boom from the fully compressed configuration to the fully \textbf{D}eployed state. As shown in Figure~\ref{fig:hybrid}(D), the force--displacement curve shows two sets of drops corresponding to different module-level transitions.
The first local drop occurs between points $i$ and $ii$, corresponding to the sudden ``unpacking'' of the bottom Yoshimura module ($\beta_3 = 35^\circ$), where stored elastic energy is released as the module transitions from the self-\textbf{P}acked (\textbf{P}) state to the \textbf{F}olded (\textbf{F}) state.
The second set of snapping events, beginning between points $iii$ and $iv$, marks the transition from \textbf{F}olded (\textbf{F}) to \textbf{D}eployed (\textbf{D}) states (i.e., reversed bucking). This snap starts when the top layer of rhombus facets pops out, releasing the energy accumulated during extension. Once this initial release occurs, the remaining layers of rhombus facets pop out sequentially from top to bottom---the same sequence as in the compression test.

\medskip
Together, the compression and tensile tests show that $\beta$ can be used as a mechanical programming parameter: smaller-$\beta$ modules have lower \textbf{D}eployed-to-\textbf{F}olded critical buckling forces, whereas larger-$\beta$ modules can access a stronger self-\textbf{P}acked state. This ordered response provides the design basis for the deployable robotic crane below, where pneumatic pressure and tendon actuation are used to move a multi-module boom among selected states.

\section{Hyper-Yoshimura Crane: Enabling Reversible and Repeatable Buckling towards Engineering Functions}

Building on the programmed sequential buckling behavior, we further render buckling \textit{reversible} and \textit{repeatable} through the combined use of tendon-driven actuation and pneumatic pressurization. The resulting hyper-Yoshimura boom can autonomously deploy and reconfigure among distinct stable configurations, enabling adaptation to diverse task requirements.
Here, inspired by concepts for future extraterrestrial-habitat construction, we demonstrate the potential of this buckling mechanism through a robotic-crane platform, termed the ``hyper-Yoshimura crane.'' Although demonstrated at a larger scale, the same architectural principle may also be applicable to smaller-scale biomedical devices and human--robot collaborative systems. The complete demonstration platform and representative tasks are summarized in \textbf{Figure~\ref{fig:crane}}.

\medskip
Following the analytical and experimental results in the previous section, we fabricate a larger-scale hyper-Yoshimura boom with six modules ($n=3, m=6$). The top three modules use the golden-ratio sector angle $\beta=31.72^\circ$, and the bottom three modules use $\beta=35^\circ$ (SI Sections 3-5, Figures S1 and S2 detail the intermediate design optimization and sequential buckling and deployment behavior of this hyper-Yoshimura boom).

\medskip
This Yoshimura boom is integrated with simple pneumatic and tendon actuation (Figure~\ref{fig:crane}A). Three tendons are routed through top and bottom tendon locators and anchored to servo motors mounted on a pulley. When activated, these motors pull the tendons and compress the hyper-Yoshimura boom, inducing compressive buckling and self-packing. In contrast, a pneumatic air tube connected to an external compressor delivers pressurized air into the boom, causing the facets to pop out sequentially toward the \textbf{D}eployed state. The combined use of internal pressurization and tendon actuation therefore enables bidirectional transitions among stable states---thus making buckling reversible and repeatable.

\subsection*{Configuration Switching with Minimal Actuation}

\begin{figure}
    \centering
    \includegraphics[width=\textwidth]{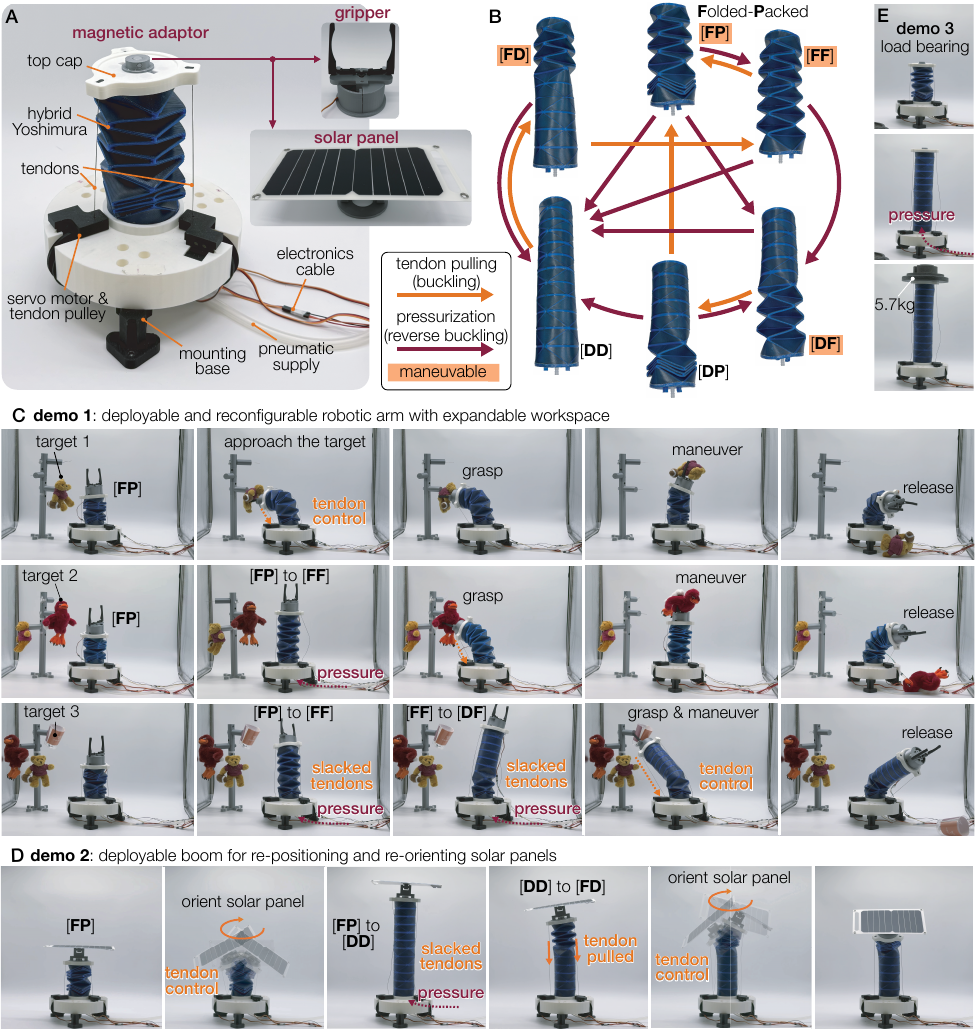}
    \caption{\textbf{Robotic crane demonstration with a hyper-Yoshimura of $\beta=31.72^\circ, 35^\circ$, $n=3$, and $m=6$.} (A) Demonstration setup highlighting pneumatic and tendon actuation and two hot-swappable end attachments: a gripper and a solar panel. (B) Switching sequence among all six possible configurations. Hyper-Yoshimura modules are soft in bending in the \textbf{F}olded states, so four configurations are maneuverable by differential tendon control. (C) Deployable manipulation demonstration with an expandable workspace. (D) Solar panel re-positioning and re-orientation demonstration. (E) Load-bearing capacity of the \textbf{D}eployed hyper-Yoshimura, which carries 28.5 times its structural weight.}
    \label{fig:crane}
\end{figure}

Mechanical testing suggests that the  Yoshimura crane boom features six stable configurations. The top three modules with $\beta=31.72^\circ$ can settle in the \textbf{F}olded (\textbf{F}) and \textbf{D}eployed (\textbf{D}) states, while the bottom three modules with $\beta=35^\circ$ can settle in \textbf{F}, \textbf{D}, and self-\textbf{P}acked (\textbf{P}) states. For clarity, we label these six stable configurations using paired \textbf{P}, \textbf{D}, and \textbf{F} acronyms. For example, ``[\textbf{FP}]'' means the top three $\beta=31.72^\circ$ modules are in the \textbf{F}olded state and the bottom three $\beta=35^\circ$ modules are in the self-\textbf{P}acked state.
Altogether, there are six symmetric stable configurations [\textbf{FP}], [\textbf{FF}], [\textbf{FD}], [\textbf{DP}], [\textbf{DF}], and [\textbf{DD}]. 

\medskip
We demonstrate that a simple combination of pneumatic and tendon actuation can switch the Yoshimura boom among all six stable configurations (Figure~\ref{fig:crane}B and Movie S8). Pneumatic pressure generally drives the boom from more compact \textbf{P} states toward extended \textbf{F} and \textbf{D} states, whereas tendon pulling drives the opposite transition. Carefully controlling the input air volume and tendon length directs the boom to desired configurations. In the configuration-switching diagram in Figure~\ref{fig:crane}(B), red arrows represent pneumatic deployment and orange arrows indicate tendon-driven compression. Some configuration switches require only one actuation input. For example, transitioning from Folded-Folded [\textbf{FF}] to Deployed-Folded [\textbf{DF}] only requires pneumatic pressure. By contrast, transitioning from [\textbf{FF}] to [\textbf{FD}] requires a larger pneumatic input to reach [\textbf{DD}], followed by tendon pulling back into [\textbf{FD}].
The transition from [\textbf{FD}] to [\textbf{FF}] is the most demanding because it must buckle the $\beta=35^\circ$ modules from the \textbf{D}eployed state toward their \textbf{F}olded state. Therefore, one must choose sufficiently strong motors to provide the necessary critical bucking force. 

\medskip
Once the hyper-Yoshimura trunk settles into a targeted configuration, differential tendon pulling can bend the flexible, \textbf{F}olded, section(s) for robotic operation. Four configurations, [\textbf{FF}], [\textbf{DF}], [\textbf{DP}], and [\textbf{FD}], are therefore maneuverable and can support task execution. The differential tendon pulling is controlled with an Arduino R4 platform, which integrates joystick input, command buttons, and a four-digit display to manage motor actuation, gripper operation, and real-time operation (control setup detailed in SI Section 5 and Figure S3).

\subsection*{Demonstration 1: Deployable Manipulation with Expandable Workspace}


To demonstrate the versatility of the hyper-Yoshimura crane as a manipulator, we attach a gripper to the boom tip and perform object-handling tasks across different working spaces (Figure~\ref{fig:crane}C). Unlike conventional manipulators with fixed length and a fixed reachable workspace, the Yoshimura crane adjusts its working range by transitioning between stable configurations. Three objects are placed at different heights, requiring the crane to adapt its length to access each target. First, the crane remains in its initial folded and packed [\textbf{FP}] configuration to grip and relocate the nearest object (Movie S3). Then, the boom transitions from [\textbf{FP}] to the elongated [\textbf{FF}] configuration through simple pressurization, expanding the reachable workspace and allowing manipulation of the second object at a higher position (Movie S4). Finally, an additional transition from [\textbf{FF}] to the \textbf{D}eployed-\textbf{F}olded [\textbf{DF}] configuration enables the crane to grasp and relocate a third object placed at an even greater height (Movie S5). After each relocation, the boom retracts back to its initial configuration through tendon actuation and pneumatic control, preparing the system for the next task. The [\textbf{DF}] configuration, despite being the most extended, is also comparatively easy to steer because the \textbf{D}eployed modules near the tip remain stiff while the flexible \textbf{F}olded modules stay near the base, allowing tendon pulling to position the gripper efficiently.
%

\subsection*{Demonstration 2: Solar Panel Re-Positioning and Re-Orientation}
Inspired by deployable and controllable solar-panel assemblies for future lunar-surface habitats \cite{tiffin2023tall, song2023sizing}, we demonstrate how the hyper-Yoshimura crane dynamically positions an attached solar panel through structural reconfiguration and tendon-driven maneuvering (Figure~\ref{fig:crane}D). In the initial Folded--Packed [\textbf{FP}] state, the crane remains sufficiently flexible to rotate the tip-mounted solar panel and orient it toward the Sun. To extend its reach in response to, for example, seasonal variations in solar position, the boom is first pneumatically driven to the fully deployed [\textbf{DD}] configuration and subsequently compressed by tendon actuation into the flexible--deployed [\textbf{FD}] state. In this [\textbf{FD}] state, the solar panel is supported at an elevated position while retaining the flexibility required for active reorientation (Movie S6). Pneumatic pressurization therefore selects among stable configurations and adjusts the panel height, whereas differential tendon actuation bends and rotates the boom to control panel orientation. Note that, once the desired configuration is reached, the folded geometry maintains the corresponding deployment height without continuous pneumatic pressurization, allowing active tendon control to be dedicated to directional adjustment. 

\subsection*{Demonstration 3: Load-Bearing Capability of the Yoshimura Crane}

The hyper-Yoshimura crane also exhibits load-bearing capability without requiring continuous pneumatic pressurization (Figure~\ref{fig:crane}E). To test this function, we transition the crane from the [\textbf{FP}] state to the fully deployed [\textbf{DD}] configuration and place a 5.7 kg weight on the top end. Although the boom weighs only 0.2 kg, it supports this end load without facet buckling, corresponding to a load-to-weight ratio of approximately 28.5. This performance exceeds typical ratios reported for 3D-printed deployable trunks or similar lightweight robotic structures~\cite{park2022deployable,amadeo2022soft}, highlighting the mechanical efficiency and practical robustness of the Yoshimura-based design (Movie S7).

\medskip
These demonstrations collectively show that the same geometry-programmed stable configurations can support manipulation, solar-panel reorientation, and load-bearing tasks. Many of the delays observed during the demonstrations are primarily due to operator-controlled timing of the input sequence rather than mechanical limitations of the structure itself. Future integration of feedback control is expected to enable smoother and more continuous operation.

\section{Conclusion}
This study introduces a new origami pattern, termed ``hyper-Yoshimura,'' by extending the decades-old Yoshimura design rule through an increase in the sector angle $\beta$. Although this geometric modification seems simple, it allows the hyper-Yoshimura structure to obtain sequential buckling behavior and additional elastically stable states, enabling autonomous deployment and reconfiguration. Specifically, increasing $\beta$ beyond the classical flat-foldability condition, $90^\circ/n < \beta < 45^\circ$, causes the kinematically admissible \textbf{F}olded state to remain extended rather than becoming fully compact. The structure can therefore undergo further compression through out-of-plane ``hyperfolding.'' This process produces a self-\textbf{P}acked state that expands the accessible set of elastically stable configurations to include \textbf{D}eployed, \textbf{F}olded, and self-\textbf{P}acked states with distinct geometries and mechanical properties.

\medskip
Bar-network modeling reveals that the sector angle $\beta$ regulates two distinct features of the \textbf{D}eployed-to-\textbf{F}olded buckling transition. As $\beta$ increases, the potential-energy barrier $\Delta U_{\mathbf{D}\rightarrow\mathbf{F}}$ decreases, whereas the critical buckling force $F_\mathrm{cr}$ increases, indicating that the energy barrier becomes narrower and steeper. Single-module experiments validate the predicted dependence of the buckling force on $\beta$. Hybrid stacking further establishes a system-level design rule: assigning distinct $\beta$ values to different modules separates their buckling thresholds, thereby transforming a global axial load into ordered sequential buckling. Thus, the single geometric parameter $\beta$ simultaneously programs the accessible stable states and the associated buckling order, or state switching sequence.

\medskip
A six-module hyper-Yoshimura crane demonstrates how this geometrically encoded buckling and state switching can be translated into engineering functionality. In this system, pneumatic pressurization drives deployment, whereas tendon actuation enables controlled compression and buckling. Differential tendon actuation further permits bending and maneuvering of the crane. Laboratory demonstrations show that the hyper-Yoshimura crane can manipulate objects within an expandable workspace, reposition and reorient solar panels mounted at its tip, and support payloads without continuous pneumatic pressurization.

\medskip
Several directions remain for future development. The discrete bar-network model can be extended to incorporate crease bending, self-contact, and finite material thickness, thereby improving predictions of the post-buckling response. Hyper-Yoshimura structures may also be explored across a broader range of length scales and material selections. In addition, increasing the number of circumferential rhombus units, $n$, may further expand the accessible performance space and functionalities. Nevertheless, the kinematic analysis, mechanical modeling, prototyping, and functional demonstrations presented here establish a comprehensive foundation for hyper-Yoshimura as a framework for programming and harnessing buckling instabilities in engineering applications.

\section{Methods}
\subsection*{Discrete Bar-Network Model}
\label{subsec:model}
To compute the elastic energy stored in the hyper-Yoshimura module as it is compressed axially from the \textbf{D}eployed to the self-\textbf{P}acked state, we model the module as a discrete bar network. The network geometry is built according to the $n=3$ Yoshimura crease pattern, comprising a base layer and a top layer of three rhombi each, with their boundary vertices interconnected through a mid-layer of vertices located at the rhombus tips (Figure~\ref{fig:single_module}A). This construction generates $6n = 18$ nodes arranged on three concentric regular triangles, connected by bars representing the mountain folds, valley folds, and virtual folds of the rhombus facets. Because every facet of the Yoshimura pattern is triangulated by these three families of folds, the bar constraints are sufficient to enforce full in-plane kinematic compatibility, and the elastic energy stored in axial bar stretching faithfully captures the energetic cost of deforming away from a kinematically admissible configuration. Since the Yoshimura pattern is not rigid-foldable, crease bending stiffness plays a comparatively minor role in shaping the potential energy landscape and is therefore omitted from the present model. Incorporating crease-bending stiffness is a natural extension that would calibrate the model's predictions more closely to physical experiments, but it would not fundamentally alter the potential energy landscape's topology.

\medskip
Each node $i$ carries mass $m_i$ and position $\mathbf{x}_i$, evolving under Newton's second law. The bars connecting node $i$ to its neighbors $\mathcal{N}(i)$ act as linear springs with stiffness $k$ and rest length $l_{ij}^0$ equal to the corresponding crease length in the undeformed Yoshimura tessellation defined in Figure~\ref{fig:overview}. The equation of motion is therefore
\begin{equation}
    m_i \ddot{\mathbf{x}}_i = -\sum_{j \in \mathcal{N}(i)} k\left(\|\mathbf{x}_i - \mathbf{x}_j\| - l_{ij}^0\right) \hat{\mathbf{r}}_{ij} - 2\zeta\sqrt{m_i k}\sum_{j \in \mathcal{N}(i)} \left(\dot{\mathbf{x}}_i - \dot{\mathbf{x}}_j\right) + \mathbf{F}_i^\text{ext},
    \label{eq:eom}
\end{equation}
where $\hat{\mathbf{r}}_{ij} = (\mathbf{x}_i - \mathbf{x}_j)/\|\mathbf{x}_i - \mathbf{x}_j\|$, $\zeta$ is the bar damping ratio, and $\mathbf{F}_i^\text{ext}$ collects external forces and any prescribed-motion constraints that drive the mass nodes. The total elastic potential energy stored in the network is the sum of the bar-stretch contributions, $U = \tfrac{1}{2} \sum_{(i,j)} k\,(\|\mathbf{x}_i - \mathbf{x}_j\| - l_{ij}^0)^2$, and the axial force conjugate to the slant height $d$ is $F = -\partial U/\partial d$.

\medskip
We integrate Eq.~(\ref{eq:eom}) with a hybrid explicit--projective scheme~\cite{phalak2026openprc}: a fourth-order Runge--Kutta step advances positions and velocities under the force law, followed by a position-based projection that enforces any rigid constraints to within a specified tolerance. In the present study, all bars are treated as soft springs and the projection stage is inactive, but the same machinery extends naturally to mixed soft--rigid networks in future work.

\medskip
The effective stiffness parameter of the reduced-order bar-network model was calibrated against the Instron force--displacement measurements for a single reference specimen. Calibration was performed manually by comparing the simulated and measured force--displacement curves and adjusting the stiffness scale to match the magnitude of the first compressive-force peak, corresponding to the onset of the \textbf{D}eployed-to-\textbf{F}olded transition. The resulting parameter set was then held fixed in all subsequent simulations, with the network geometry and bar rest lengths updated according to the prescribed sector angle $\beta$. Thus, the force scale was calibrated using one experimental case, while the responses at the remaining sector angles were evaluated without further parameter adjustment.

\subsection*{3D Printing and Assembly of hyper-Yoshimura}
Hyper-Yoshimura specimens were fabricated as flat panels by dual-material fused deposition modeling using an Ultimaker S5 printer. Ultimaker TPU 95A formed the compliant substrate and crease regions, while Ultimaker Nylon provided localized stiffness in the triangular facets. For the pneumatic boom, the panels were coated to improve airtightness, bonded using tetrahydrofuran with reinforcing TPU strips, and assembled into a tubular structure by welding. The boom was thermally conditioned at $120\,^\circ\mathrm{C}$ for 2~hours, followed by brief shape-setting treatments at $140\,^\circ\mathrm{C}$ and end-cap installation. More detailed coating, assembly, and annealing procedures are provided in Section~4 of the Supplementary Material.

\medskip
\textbf{Supporting Information} \par
Supporting Information is available from the Wiley Online Library or from the author.

\medskip
\textbf{Acknowledgements} \par 
Z.Z, Y.P., V.D. and S.L. acknowledge support from the National Science Foundation (CMMI-2312422) and Virginia Tech (via startup and graduate assistantship).  I.W. acknowledges support from the National Science Foundation (CMMI-2506466)

\medskip

\bibliographystyle{ieeetr}
\bibliography{reference}

@article{deshpande2024golden,
  title={‘Golden Ratio Yoshimura’for meta-stable and massively reconfigurable deployment},
  author={Deshpande, Vishrut and Phalak, Yogesh and Zhou, Ziyang and Walker, Ian and Li, Suyi},
  journal={Philosophical Transactions of the Royal Society A: Mathematical, Physical and Engineering Sciences},
  volume={382},
  number={2283},
  year={2024},
  publisher={The Royal Society}
}

@article{phalak2026openprc,
  title={OpenPRC: A Unified Open-Source Framework for Physics-to-Task Evaluation in Physical Reservoir Computing},
  author={Phalak, Yogesh and Lor, Wen Sin and Khairnar, Apoorva and Jantzen, Benjamin and Naughton, Noel and Li, Suyi},
  journal={arXiv preprint arXiv:2604.07423},
  year={2026}
}

@incollection{tiffin2023tall,
  title={Tall Lunar Towers: Systems Analysis of Lunar-Surface-Assembled Power, Communication, and Navigation Infrastructure},
  author={Tiffin, Daniel and Mahlin, Matthew},
  booktitle={ASCEND 2023},
  pages={4754},
  year={2023}
}

@inproceedings{song2023sizing,
  title={Sizing and design tool for tall lunar tower},
  author={Song, Kyongchan and Mikulas, Martin and Mahlin, Matthew K and Cassady, Jacob T},
  booktitle={AIAA SciTech 2023 Forum},
  pages={0382},
  year={2023}
}

@article{park2022deployable,
  title={Deployable soft origami modular robotic arm with variable stiffness using facet buckling},
  author={Park, MinJo and Kim, Woongbae and Yu, Sung-Yol and Cho, Jungmin and Kang, Wonkyeong and Byun, Junghwan and Jeong, Useok and Cho, Kyu-Jin},
  journal={IEEE Robotics and Automation Letters},
  volume={8},
  number={2},
  pages={864--871},
  year={2022},
  publisher={IEEE}
}

@article{amadeo2022soft,
  title={Soft robotic deployable origami actuators for neurosurgical brain retraction},
  author={Amadeo, Tomas and Van Lewen, Daniel and Janke, Taylor and Ranzani, Tommaso and Devaiah, Anand and Upadhyay, Urvashi and Russo, Sheila},
  journal={Frontiers in Robotics and AI},
  volume={8},
  pages={731010},
  year={2022},
  publisher={Frontiers Media SA}
}

@article{bertoldi2017flexible,
  title={Flexible mechanical metamaterials},
  author={Bertoldi, Katia and Vitelli, Vincenzo and Christensen, Johan and Van Hecke, Martin},
  journal={Nature Reviews Materials},
  volume={2},
  number={11},
  pages={1--11},
  year={2017},
  publisher={Nature Publishing Group}
}

@article{kong2026origami,
  title={From Origami to Bistable and Laminate Structures: A Review for Multifunctional Applications from Structural Perspective of Shape-Changing Structures},
  author={Kong, Lingchen and Zhao, Yaoyao Fiona},
  journal={Advanced Intelligent Systems},
  volume={8},
  number={2},
  pages={2500505},
  year={2026},
  publisher={Wiley Online Library}
}

@article{yan2025snap,
  title={Snap-through instability in mechanical metamaterials},
  author={Yan, Sen and Wu, Lingling and Wen, Yongzheng and Sun, Jingbo and Zhou, Ji},
  journal={Responsive Materials},
  volume={3},
  number={1},
  pages={e20240035},
  year={2025},
  publisher={Wiley Online Library}
}

@article{zhou2025multistable,
  title={Multistable thin-shell metastructures for multiresponsive reconfigurable metabots},
  author={Zhou, Caizhi and Qing, Haitao and Sun, Haoze and Qi, Fangjie and Hong, Yaoye and Yin, Jie},
  journal={Science Advances},
  volume={11},
  number={42},
  pages={eadx4359},
  year={2025},
  publisher={American Association for the Advancement of Science}
}

@article{levine2021materials,
  title={Materials with electroprogrammable stiffness},
  author={Levine, David J and Turner, Kevin T and Pikul, James H},
  journal={Advanced Materials},
  volume={33},
  number={35},
  pages={2007952},
  year={2021},
  publisher={Wiley Online Library}
}

@article{gao2022stiffness,
  title={Stiffness variable polymers comprising phase-changing side-chains: material syntheses and application explorations},
  author={Gao, Meng and Meng, Yuan and Shen, Claire and Pei, Qibing},
  journal={Advanced Materials},
  volume={34},
  number={21},
  pages={2109798},
  year={2022},
  publisher={Wiley Online Library}
}

@article{jiao2023mechanical,
  title={Mechanical metamaterials and beyond},
  author={Jiao, Pengcheng and Mueller, Jochen and Raney, Jordan R and Zheng, Xiaoyu and Alavi, Amir H},
  journal={Nature communications},
  volume={14},
  number={1},
  pages={6004},
  year={2023},
  publisher={Nature Publishing Group UK London}
}

@article{chen2021reprogrammable,
  title={A reprogrammable mechanical metamaterial with stable memory},
  author={Chen, Tian and Pauly, Mark and Reis, Pedro M},
  journal={Nature},
  volume={589},
  number={7842},
  pages={386--390},
  year={2021},
  publisher={Nature Publishing Group UK London}
}

@article{zaiser2023disordered,
  title={Disordered mechanical metamaterials},
  author={Zaiser, Michael and Zapperi, Stefano},
  journal={Nature Reviews Physics},
  volume={5},
  number={11},
  pages={679--688},
  year={2023},
  publisher={Nature Publishing Group UK London}
}

@article{sun2026magnetic,
  title={Magnetic coupling transforms random snapping into ordered sequences in soft metamaterials},
  author={Sun, Haoze and Alkuino, Gabriel and Chi, Yinding and Zabila, Yevhen and Qing, Haitao and Makarov, Denys and Zhang, Teng and Yin, Jie},
  journal={Science Advances},
  volume={12},
  number={12},
  pages={eaec3182},
  year={2026},
  publisher={American Association for the Advancement of Science}
}

@article{zhou2024self,
  title={Self-Sustained and Coordinated Rhythmic Deformations with SMA for Controller-Free Locomotion},
  author={Zhou, Ziyang and Li, Suyi},
  journal={Advanced Intelligent Systems},
  volume={6},
  number={5},
  pages={2300667},
  year={2024},
  publisher={Wiley Online Library}
}

@article{djellouli2024shell,
  title={Shell buckling for programmable metafluids},
  author={Djellouli, Adel and Van Raemdonck, Bert and Wang, Yang and Yang, Yi and Caillaud, Anthony and Weitz, David and Rubinstein, Shmuel and Gorissen, Benjamin and Bertoldi, Katia},
  journal={Nature},
  volume={628},
  number={8008},
  pages={545--550},
  year={2024},
  publisher={Nature Publishing Group UK London}
}

@article{luo2024intrinsically,
  title={Intrinsically multistable soft actuator driven by mixed-mode snap-through instabilities},
  author={Luo, Yichi and Patel, Dinesh K and Li, Zefang and Hu, Yafeng and Luo, Hao and Yao, Lining and Majidi, Carmel},
  journal={Advanced Science},
  volume={11},
  number={18},
  pages={2307391},
  year={2024},
  publisher={Wiley Online Library}
}

@article{bichara2023multi,
  title={A multi-stable deployable quadrifilar helix antenna with radiation reconfigurability for disaster-prone areas},
  author={Bichara, Rosette Maria and Costantine, Joseph and Tawk, Youssef and Sakovsky, Maria},
  journal={Nature Communications},
  volume={14},
  number={1},
  pages={8511},
  year={2023},
  publisher={Nature Publishing Group UK London}
}

@article{yang2024kirigami,
  title={Kirigami electronics for long-term electrophysiological recording of human neural organoids and assembloids},
  author={Yang, Xiao and Forr{\'o}, Csaba and Li, Thomas L and Miura, Yuki and Zaluska, Tomasz J and Tsai, Ching-Ting and Kanton, Sabina and McQueen, James P and Chen, Xiaoyu and Mollo, Valentina and others},
  journal={Nature biotechnology},
  volume={42},
  number={12},
  pages={1836--1843},
  year={2024},
  publisher={Nature Publishing Group US New York}
}

@article{lamoureux2015dynamic,
  title={Dynamic kirigami structures for integrated solar tracking},
  author={Lamoureux, Aaron and Lee, Kyusang and Shlian, Matthew and Forrest, Stephen R and Shtein, Max},
  journal={Nature communications},
  volume={6},
  number={1},
  pages={8092},
  year={2015},
  publisher={Nature Publishing Group UK London}
}

@article{lu2023origami,
  title={Origami with rotational symmetry: a review on their mechanics and design},
  author={Lu, Lu and Leanza, Sophie and Zhao, Ruike Renee},
  journal={Applied Mechanics Reviews},
  volume={75},
  number={5},
  pages={050801},
  year={2023},
  publisher={American Society of Mechanical Engineers}
}

@article{misseroni2024origami,
  title={Origami engineering},
  author={Misseroni, Diego and Pratapa, Phanisri P and Liu, Ke and Kresling, Biruta and Chen, Yan and Daraio, Chiara and Paulino, Glaucio H},
  journal={Nature Reviews Methods Primers},
  volume={4},
  number={1},
  pages={40},
  year={2024},
  publisher={Nature Publishing Group UK London}
}

@article{zhang2023plug,
  title={Plug \& play origami modules with all-purpose deformation modes},
  author={Zhang, Chao and Zhang, Zhuang and Peng, Yun and Zhang, Yanlin and An, Siqi and Wang, Yunjie and Zhai, Zirui and Xu, Yan and Jiang, Hanqing},
  journal={Nature communications},
  volume={14},
  number={1},
  pages={4329},
  year={2023},
  publisher={Nature Publishing Group UK London}
}

@article{ze2022soft,
  title={Soft robotic origami crawler},
  author={Ze, Qiji and Wu, Shuai and Nishikawa, Jun and Dai, Jize and Sun, Yue and Leanza, Sophie and Zemelka, Cole and Novelino, Larissa S and Paulino, Glaucio H and Zhao, Ruike Renee},
  journal={Science advances},
  volume={8},
  number={13},
  pages={eabm7834},
  year={2022},
  publisher={American Association for the Advancement of Science}
}

@article{schenk2013geometry,
  title={Geometry of Miura-folded metamaterials},
  author={Schenk, Mark and Guest, Simon D},
  journal={Proceedings of the National Academy of Sciences},
  volume={110},
  number={9},
  pages={3276--3281},
  year={2013},
  publisher={National Academy of Sciences}
}

@article{wang2023building,
  title={Building intelligence in the mechanical domain—Harvesting the reservoir computing power in origami to achieve information perception tasks},
  author={Wang, Jun and Li, Suyi},
  journal={Advanced Intelligent Systems},
  volume={5},
  number={9},
  pages={2300086},
  year={2023},
  publisher={Wiley Online Library}
}

@article{wang2026embodying,
  title={Embodying physical computing into soft robots},
  author={Wang, Jun and Zhou, Ziyang and Kahak, Ardalan and Li, Suyi},
  journal={Nature Communications},
  volume={17},
  number={1},
  pages={2455},
  year={2026},
  publisher={Nature Publishing Group UK London}
}

@article{boron2026stochastic,
  title={Stochastic Entanglement of Deterministic Origami Tentacles For Robust Robotic Grasping},
  author={Boron, Alec and Zheng, Bokun and Zhou, Ziyang and Naughton, Noel and Li, Suyi},
  journal={Advanced Science},
  pages={e76810},
  year={2026},
  publisher={Wiley Online Library}
}

@article{wang2026coupled,
  title={Coupled effects of nanoparticle surface roughness and grafting architecture on mechanical reinforcement in polymer nanocomposites},
  author={Wang, Haoyu and Yang, Zhangke and Meng, Zhaoxu},
  journal={Nanotechnology},
  volume={37},
  number={35},
  pages={355702},
  year={2026},
  publisher={IOP Publishing}
}

\newpage
\begin{table}[htbp]
\centering
\scriptsize

\renewcommand{\arraystretch}{2}
\setlength{\tabcolsep}{3.5pt} 
\begin{tabular}{|c||c|c|c|c|c|}
\hline
\textbf{States}
& $\beta$
& $\theta$
& $d$
& $\chi$
& $c=1-d/\tan\beta$ \\
\hline\hline

Deployed
& $(0^\circ,90^\circ)$
& $180^\circ$
& $w$
& $180^\circ$
& $0$ \\

Folded
& $\left[\frac{90^\circ}{n},90^\circ\right)$
& $\theta_\mathbf{F}$
& $d_\mathbf{F}$
& $180^\circ$
& $1-\sqrt{1-\frac{\tan^2(90^\circ/n)}{\tan^2\beta}}$ \\

Self-Packed
& $\left(\frac{90^\circ}{n},45^\circ\right)$
& $\theta_\mathbf{F}$
& $0$
& $\cos^{-1}\!\left(
1-\frac{2\sin^2(90^\circ/n)}{\sin^2\beta}
\right)$
& $1$ \\

\hline
\end{tabular}
\caption{Summary of the geometric conditions for the \textbf{D}eployed, \textbf{F}olded, and self-\textbf{P}acked states, under the non-dimensional convention $l=1$.  At the \textbf{F}olded state, $d_\mathbf{F}=\sqrt{\tan^2\beta-\tan^2(90^\circ/n)}$ and $\theta_\mathbf{F}=2\sin^{-1}(d_\mathbf{F}/w)=2\cos^{-1}[\tan(90^\circ/n)/\tan\beta]$. All angular variables follow the included-dihedral convention: $\theta$ applies to mountain and valley creases, whereas $\chi$ applies to virtual creases. An unbent crease therefore has an angle of $180^\circ$. During hyperfolding, the original crease dihedrals remain at $\theta_\mathbf{F}$ while $\chi$ decreases below $180^\circ$. A distinct self-\textbf{P}acked state exists only for $90^\circ/n<\beta<45^\circ$; at the lower boundary it coincides with the flat-Folded state, and at the upper boundary the facets reach the self-intersection limit. The normalized compression ratio is $c=1-d/\tan\beta$, with $c=0$ in the Deployed state and $c=1$ in the self-Packed state; at the classical flat-foldability boundary, the Folded state also reaches $c=1$.}
\label{tab:states_parameters}
\end{table}

\end{document}